\pdfoutput=1

\documentclass[11pt]{article}

\usepackage[review]{acl}

\usepackage{times}
\usepackage{latexsym}

\usepackage[T2A,T1]{fontenc}

\usepackage[utf8]{inputenc}

\usepackage{microtype}

\usepackage{inconsolata}

\usepackage{algorithm} 
\usepackage{algpseudocode}
\usepackage{amsmath}
\usepackage{booktabs}
\usepackage{graphicx}
\usepackage{stmaryrd}
\usepackage{tipa}

\title{S{\~o}najaht: {D}efinition {E}mbeddings and {S}emantic {S}earch for {R}everse {D}ictionary {C}reation}

\author{Aleksei Dorkin \and Kairit Sirts \\
        Institute of Computer Science \\
        University of Tartu \\
        \texttt{\{aleksei.dorkin, kairit.sirts\}@ut.ee} \\ }

\begin{document}
\maketitle
\begin{abstract}

We present an information retrieval based reverse dictionary system using modern pre-trained language models and approximate nearest neighbors search algorithms. The proposed approach is applied to an existing Estonian language lexicon resource, Sõnaveeb (\textit{word web}), with the purpose of enhancing and enriching it by introducing cross-lingual reverse dictionary functionality powered by semantic search.
The performance of the system is evaluated using both an existing labeled English dataset of words and definitions that is extended to contain also Estonian and Russian translations, and a novel unlabeled evaluation approach that extracts the evaluation data from the lexicon resource itself using synonymy relations.
Evaluation results indicate that the information retrieval based semantic search approach without any model training is feasible, producing median rank of 1 in the monolingual setting and median rank of 2 in the cross-lingual setting using the unlabeled evaluation approach, with models trained for cross-lingual retrieval and including Estonian in their training data showing superior performance in our particular task.

\end{abstract}

\section{Introduction}

A reverse dictionary (see examples in Table~\ref{tab:exemples}) is a system that takes user descriptions or definitions as input and returns words or expressions corresponding to the provided input~\cite{hill2016learning, bilac2004dictionary}. The usefulness of a reverse dictionary is multi-faceted. 
It can help resolve the tip of the tongue problem---a common cognitive experience where a person is unable to recall a familiar word, despite feeling that they know it and that it is just on the verge of being remembered~\cite{brown1966tip}. For writers, it can be helpful, similarly to a thesaurus, in making the vocabulary in their work richer and more expressive. Finally, in a cross-lingual setting, the reverse dictionary allows language learners to look up words simply by describing them in their native language. Consider, for example, the accidental gap in semantics---a situation when a certain concept expressed by a word in one language does not have such an expression in another language, thus making a direct translation impossible. In this case, describing the concept represented by the word might be sufficient to find related concepts in the other language.

Early approaches to building reverse dictionary systems were based on information retrieval (IR) techniques reliant on exact term matching: both user inputs and candidate collections were represented using sets of keywords or sparse term-based vector representations~\cite{bilac2004dictionary, shaw2011building}. Such representations are very limited in their ability to represent the compositional meaning of sentences due to texts being represented as simple collections of discrete terms. 
More recent works on reverse dictionary focused on training models to reconstruct word embeddings \cite{hill2016learning,zhang2020multi} or on fine-tuning pre-trained transformers \cite{yan2020bert,tsukagoshi2021defsent,mane2022wordalchemy}. 
The main limitation of these approaches is that they require labeled data to train predictive models and, as such, are not easily generalizable to new settings or languages.

 \begin{table}[ht]
    \centering
    \begin{tabular}{cll}
    \toprule
    \multicolumn{3}{p{7cm}}{Query: \textit{``tugev emotsionaalne füüsiline või vaimne külgetõmme kellegi suhtes''}} \\
    \midrule
    \multicolumn{3}{p{7cm}}{Translation: \textit{``a strong emotional feeling of physical or mental attraction towards somebody''}} \\
    \midrule
    Rank & Word & Translation \\
    1 & armastama & to love \\
    3 & armastaja & lover \\
    4 & \textbf{armastus} & love \\
    6 & armupalang & love fervour\\
    9 & armunud & in love \\
    \midrule

    \multicolumn{3}{p{7cm}}{Query:  {\fontencoding{T2A}\selectfont \textit{``Группа людей, таких как мать-отец и дети, которые все родственники''}} } \\
    \midrule
    \multicolumn{3}{p{7cm}}{Translation: \textit{``a group of people like a mother, father and children who are all related''}} \\
    \midrule

    Rank & Word & Translation \\
    1 & abielu & marriage\\
    2 & asurkond & population \\
    27 & kollektiiv & group \\
    46 & noorpere & young family \\
    51 & \textbf{pere}& family \\
    \midrule
    \multicolumn{3}{p{7cm}}{Query: 
    \textit{``when you tell other people that something is very good and the right choice''}
    } \\
    \midrule
    Rank & Word & Translation \\
    1 & austama & to respect \\
    13 & jaatama & to agree \\
    44 & meelitama & to convince \\
    76 & \textbf{soovitama} & to recommend \\
    98 & ülistama & to praise \\
    \bottomrule
    \end{tabular}
    \caption{Examples of the reverse dictionary search. The target words are in Estonian, while the query can be in different languages. The target word is marked in bold.}
    \label{tab:exemples}
\end{table}

While the earlier IR-based approaches were limited by the expressive power of sparse text vectors and term-based text representations, dense sentence representations of modern pre-trained transformer-based language models make these problems obsolete and provide suitable representations for implementing semantic search functionality \citep{muennighoff2022sgpt}.
When applied to lexicographical data, semantic search may be leveraged to create a reverse dictionary system. Word definitions encoded by a pre-trained language model represent the search index, which is then queried with the encoded representation of the user's input (definition or description of a concept).

In this work, we develop an IR-based reverse dictionary system implementing semantic search via pre-trained transformer language model representations. 
We apply and evaluate the system on an existing Estonian linguistic resource Sõnaveeb,\footnote{\url{https://sonaveeb.ee/?lang=en}} calling the extended reverse dictionary functionality Sõnajaht (\textit{word hunt}). Sõnaveeb is the Estonian language portal of the Institute of the Estonian Language (EKI), giving public access to several lexicons.\footnote{In this work we experiment specifically with the ``Ühendsõnastik 2023'' lexicon---the combined dictionary.} A user can query the Sõnaveeb with words in several languages, such as Estonian or English. The words used for querying may also be in an inflected form. However, the current system does not support approximate search---the user has to spell the words precisely. Search over definitions is not currently supported in any capacity.

The system we propose is based on word definitions: every word in the Sõnaveeb has at least one distinct sense, and each sense has at least one definition. Any given definition in the Sõnaveeb can be linked back to its corresponding word and to a specific sense of that word. We encode the definitions using a pre-trained language model and then store these definitions in a vector database. Then, when a user inputs their description of a desired meaning, it is encoded with the same language model to be used as a query. The approximate nearest neighbor search is then used to query the vector database to return definitions linked to corresponding words.
Although all components of this system---dense sentence representations, similarity-based search, and approximate nearest neighbors---are well-known, their combination to build a reverse dictionary functionality has, according to our knowledge, not been studied previously.

For evaluating our reverse dictionary system, we introduce a novel unlabeled evaluation approach that relies on the word relation structure present in the Sõnaveeb dictionary itself.
Additionally, we utilize and extend the labeled English dataset of words and definitions introduced by \citet{hill2016learning} by translating the target words to Estonian, as well as introducing definitions in Estonian and Russian in addition to English.

To summarize, our contributions in this work are as follows:

\begin{enumerate}
    \item A novel approach to build reverse dictionary systems combining information retrieval techniques, modern pre-trained language models, and approximate nearest neighbor search algorithms;
    \item A novel unlabeled evaluation approach intended to gauge the performance of a given language model in the context of a reverse dictionary that does not require annotated data;
    \item An extension of an existing English reverse dictionary evaluation dataset to a cross-lingual setting by adding words and definitions in Estonian and Russian;
    \item Evaluation of a number of different pre-trained language models for their suitability for both monolingual and cross-lingual IR-based reverse dictionary task in a non-English language (Estonian);
    \item Demonstrating the utility of building an IR-based reverse dictionary system by applying it to an existing Estonian language resource.
\end{enumerate}

We make the code and data available on GitHub\footnote{\url{https://github.com/slowwavesleep/sonajaht}} and HuggingFace Hub\footnote{\url{https://huggingface.co/datasets/adorkin/sonajaht}}, respectively.

\section{Related Work}
The approaches used to address the reverse dictionary problem can be divided into two---prediction-based methods and information retrieval (IR) based methods. Both approaches assume a dictionary dataset but use it differently---while prediction-based methods use the data for training a predictive model, IR-based methods require access to a dictionary during inference.

\subsection{Prediction-based approaches} 
Prediction-based approaches have mostly framed the reverse dictionary problem as word embedding reconstruction where the model is trained to predict target word embeddings from their definition embeddings \citep{hill2016learning}.
The search is performed in two steps: first, the definition is embedded into a Word2Vec \cite{mikolov2013efficient} space, and the trained model is used to predict the target word vector. Then, the produced vector is used to search for similar vectors in the Word2Vec model's vocabulary, and the most similar entries corresponding to these vectors are returned. 

\citet{zhang2020multi} expanded on \citet{hill2016learning} by introducing additional objectives to the model, namely part-of-speech, morpheme, word category, and sememe predictors that are then used to re-score the final output.  
WantWords \cite{qi2020wantwords} adds a web interface on top of \citet{zhang2020multi} and introduces the Chinese language to the system, making both monolingual and cross-lingual searches possible.

More recent approaches have leveraged pre-trained transformer models. \citet{yan2020bert} used BERT \cite{devlin2018bert} to predict the target word as a masked sequence in the context of its definition. \citet{tsukagoshi2021defsent} fine-tuned a BERT-based classifier to predict the target word from its definition representation.
\citet{mane2022wordalchemy} adopted the encoder-decoder T5 model \cite{raffel2020exploring} to generate the word given the definition. 
\citet{tsukagoshi2021defsent} proposed a type of sentence embedding model that is trained to predict a word out of a predefined vocabulary given the definition of that word. 
Meanwhile, \citet{jo-2023-self} aimed to improve the ability of BERT to represent the semantics of short or single-word sentences by means of minimizing the distance between isolated words and their human-written definitions, as well as definitions and the words appearing in the relevant context.

\subsection{IR-based approaches}
IR-based solutions to the reverse dictionary problem assume the presence of a dictionary that contains words with their definitions. Both the user input and target word definitions are represented as vectors that are compared with some similarity measure, and the words with the most similar definition representations to the user input are returned.

Previous IR-based works fall into the pre-neural times, using count-based representations such as keyword sets and tf-idf \citep{bilac2004dictionary}.
Because the count-based representations rely on term overlap between the user input and target definitions, other works explored various heuristics to augment the representations to increase the term overlap. For instance, \citet{shaw2011building} expanded user queries with WordNet relations and reranked outputs by assigning differential weights to words according to their syntactic function in the sentence.
We are unaware of any previous work using dense vector representations for IR-based reverse dictionary search.

\section{Methodology}

Our approach to the reverse dictionary problem is, similarly to \citet{bilac2004dictionary} and \citet{shaw2011building}, based on information retrieval techniques. We assume the presence of a lexicon of words with their definitions. The system treats individual definitions as candidates in the search database, compares the user input to all candidates, and outputs the most relevant results based on cosine similarity. When a search database has a significant number of entries, a brute-force all-to-all comparison becomes computationally infeasible. Thus, we adopt the approximate nearest neighbors algorithm that reduces the computational complexity of vector search. We chose the Qdrant vector database\footnote{\url{https://qdrant.tech/}} that implements the  Hierarchical Navigable
Small World (HNSW) approximate nearest neighbors algorithm \citep{malkov2018efficient}. In our experiments, we estimate the nearest neighbors search to be approximately 60 times faster than the brute-force search. The vectors are stored in the database together with some additional metadata, such as the language of the definition and both word and definition identifiers for the ease of later filtering. To store supplementary information, such as definitions themselves and synonymy relations, we opted to use SQLite for our simple implementation.

\subsection{Database}
The source of the data is the public API of the Estonian language portal Sõnaveeb.
Due to the lack of filtering options in the API, we had to request all available information for every word entry for further processing. Out of that data, we extracted words and word definitions (represented as both surface forms and identifiers), as well as the language of words and definitions and synonymy relations for each word. We filtered the data based on the language of the words to 
keep only the words in Estonian. We kept the definitions in all available languages to evaluate the cross-lingual functionality. Synonymy relations came in several types: word-to-word, sense-to-word, and sense-to-sense relations. However only the coarse-grained word-to-word synonymy links were reliably present; thus we opted to keep only that type. Additionally, we discovered that a significant number of synonyms in the database had only a single direction from word A to word B, but not from word B to word A. We understand synonymy as a symmetrical relation; thus, for the purposes of evaluation, we mirrored every single direction synonymy. The statistics of the final dataset are shown in Table~\ref{tab:statistics}.

\begin{table}[t]
    \centering
    \setlength{\tabcolsep}{3.4pt}
    \begin{tabular}{lrr}
    \toprule
    & Number of  \\
    \midrule
    Words & 124K \\
    Definitions in Estonian & 213K  \\
    Definitions in other languages & 16K \\
    Synonyms & 295K \\
    Mirrored synonyms & 590K \\
    Synonyms per word on average & 3.85 \\
    \bottomrule
    \end{tabular}
    \caption{Statistics of the dataset extracted from the Estonian lexicon Sõnaveeb.
    } 
    \label{tab:statistics}
\end{table}

\subsection{Ground Truth}\label{gt}
One of the challenges in estimating the quality of a reverse dictionary without user feedback lies in the requirement of annotated evaluation data. Commonly, annotated data for this purpose comprises definition/target word pairs~\cite{hill2016learning}. This approach is quite limiting because one can often find several suitable words fitting a given definition, all of which can be considered correct answers.
To alleviate these issues, we propose a novel approach to defining the ground truth for reverse dictionary evaluation based on the synonymy relations of the underlying lexicon.

In our approach, we consider both the target word and its synonyms as the ground truth based on the assumption that synonymous words relate to approximately the same concepts. Thus, from the user's point of view, synonyms should be expected in the system's output in addition to the target word itself. This way, we resolve the problem of the single target word limitedness. While this approach allows us to sidestep the need for annotated data, it introduces the requirement of knowing synonymy relations between words. 
However, we do not consider this requirement too limiting because dictionaries generally contain information on synonymy relations between words. Alternatively, the synonymy relations can be extracted from other sources, such as WordNet. This approach makes it possible to use the dictionary as the source of both queries, candidates, and the ground truth.

\subsection{Evaluation settings}

We evaluate our reverse dictionary system in two settings, called \emph{unlabeled} and \emph{labeled} evaluation, respectively. In both cases, the ground truth is as described in Section~\ref{gt}. In the unlabeled case, the queries are also extracted from the Sõnaveeb dictionary, while in the labeled case, they are taken from an existing annotated dataset.

\paragraph{Unlabeled evaluation} In this setup, we have no predefined query/target word pairs. However, we have dictionary entries that map words to definitions and definitions to words. We also have information on synonymy relations from the dictionary. Thus, in unlabeled evaluation we consider every individual definition entry as a query and perform vector search over all definition vectors. When computing evaluation metrics, we only consider the queried definitions of words that have multiple definitions or synonyms, i.e., they have an associated ground truth definition other than the queried definition itself.

Unlabeled evaluation aims to compare different embedding models without requiring labeled evaluation data, which is very relevant for non-English and multilingual dictionaries. In addition, it allows the measurement of cross-lingual search capabilities of the embedding models. The algorithm is described in the pseudocode block below.

\begin{algorithm}[H]
\captionsetup{font=scriptsize}
\caption{Candidate Quality Estimation}
\label{algorithm:candidate_quality_estimation}
\footnotesize
\begin{algorithmic}[1]
\Require Database of definition vectors
\Require $N$ (number of candidates to retrieve)
\State Initialize an empty mapping $candidates$

\For{each definition ID $D$ and definition vector $V$ in the database}
    \State Extract the word ID $W$ corresponding to $V$
    \If{$W$ has associated ground truth}
        \State Use $V$ as a query to the database
        \State Retrieve the top $N$ candidates and store them in $candidates$ under key $D$
    \EndIf
\EndFor
\State Assess the quality of $candidates$ using the ground truth
\end{algorithmic}
\end{algorithm}

\paragraph{Labeled evaluation} 

We also adopt labeled evaluation using an annotated dataset to verify the validity of the unlabeled approach described above.
For this, we adapted and extended the dataset comprised of human-crafted definitions from \citet{hill2016learning}. The original dataset contains 200 words together with their definitions. 

Labeled evaluation aims to model the experience of a language learner who would attempt to search for words using descriptions in their native language or an intermediary language. The Estonian language is most commonly studied either in Russian (by Russian native speakers) or in English (either by native speakers or people with some other native language). We manually translated the words from English to Estonian and adapted the English definitions when necessary.
Then, we linked the Estonian words to their respective word senses in our database.
Finally, we translated English definitions to Estonian and Russian using machine translation.\footnote{\url{https://translate.ut.ee/}} The evaluation approach is also modified compared to \citet{hill2016learning}.
Similarly to the unlabeled approach, we use the definitions from the dataset as queries. However, we consider only the relevant word sense as the target and not all senses of that word.
Overall, the ground truth and evaluation principles are similar to the unlabeled case, making the metrics of both approaches comparable.

\subsection{Embedding Models}

We evaluated several pre-trained transformer-based models to understand their suitability for producing representations for reverse dictionary search.

\paragraph{E5} \textbf{E}mb\textbf{E}ddings from bidir\textbf{E}ctional \textbf{E}ncoder r\textbf{E}presentations \cite{wang2022text} was the highest scoring open-source multilingual embedding model in the \textit{Overall} ranking on the MTEB leaderboard \cite{muennighoff2022mteb} at the time of writing. E5 is a BERT-based bi-encoder asymmetrical retrieval model: training examples were prepended with query and passage prefixes. Thus, in our experiments, we tested E5 in two distinct environments: we encoded candidate definitions with query or passage prefixes, and in both cases, we used query-prefixed queries. 

\paragraph{LaBSE} Language-agnostic BERT Sentence Embedding \cite{feng2020language} is a multilingual sentence embedding model. It is an extension of the BERT architecture designed to generate high-quality fixed-size representations for sentences or short texts. LaBSE supports over 100 languages, including Estonian. 
At the time of writing, LaBSE takes the top position in \textit{Bi-Text Mining} category on the MTEB leaderboard \cite{muennighoff2022mteb}.

\paragraph{OpenAI} In addition to open-source models, we used the proprietary OpenAI's \textit{text-embedding-ada-002} embedding model for comparison. According to our knowledge, OpenAI has not disclosed much information on the configuration and properties of this model.

\paragraph{DistilUSE V1 and V2} are DistilBERT models trained on 15 languages (but no Estonian) and 50 languages (with Estonian), respectively, and distilled from the mUSE model (Multilingual Universal Sentence Encoder)~\cite{yang2019multilingual}, provided as part of the Sentence Transformers library \cite{reimers-2019-sentence-bert, reimers-2020-multilingual-sentence-bert}.

\paragraph{BGE} \cite{bge_embedding} was the highest-scoring open-source English-only retrieval embedding model on the MTEB leaderboard \cite{muennighoff2022mteb} at the time of writing. The reason for adding this model is to understand how well a monolingual retrieval model can generalize in a cross-lingual retrieval task.

\paragraph{MPNet} Masked and Permuted Pre-training for Language Understanding \cite{song2020mpnet} is a version of the BERT \cite{devlin2018bert} model with a different training objective. The specific model we used was additionally fine-tuned on the sentence similarity task. Similarly to BGE, the intent is to test cross-lingual generalization capabilities.

\paragraph{XLM-RoBERTa} is a state-of-the-art pre-trained language model \cite{conneau2019unsupervised} that combines the RoBERTa \cite{liu2019roberta} architecture with cross-lingual learning techniques. RoBERTa is an extension of the BERT \cite{devlin2018bert} architecture and is designed for various natural language understanding tasks. XLM-RoBERTa was trained on data with some Estonian texts in it. We used the large version in our experiments. The aim of including XLM-RoBERTa is to understand how well a multilingual non-retrieval model would perform in our task.

\paragraph{Word2Vec} Finally, a Word2Vec \cite{mikolov2013efficient} model trained on Estonian data was used as a baseline.\footnote{\url{https://github.com/estnltk/word2vec-models}} We employed a very simple approach to extract definition embeddings with the Word2Vec: definitions were tokenized based on the whitespace character, then we simply ignored tokens not present in the model's vocabulary and averaged the rest.

\subsection{Metrics}
\label{sec:metrics}

To assess the quality of each model, we employed metrics traditionally used in information retrieval works \cite{buttcher2016information}, as well as some metrics from the related works \cite{hill2016learning}.
We limit the number of items the search system outputs to \textbf{100} items. 
In the following, let \textit{Res} be the collection of retrieved items, and \textit{Rel} the set of relevant items. \textit{Res[1..k]} consists of the top k items returned by the system.

\paragraph{Precision@k} Precision at k is meant to model a user's satisfaction when presented with a ranked list of results given the query. The expectation is that the user examines every item out of k in an arbitrary order \cite{buttcher2016information}. In the case of a reverse dictionary system, this is a very reasonable expectation because the items comprise words and their usually short definitions. The $P@k$ for a single query is:

$$P@k = \frac{|\text{Res}[1..k] \cap \text{Rel}|}{k}$$

\noindent
We report the mean $P@k$ over all queries and denote it as $MP@k$.

$$MP@k = \frac{1}{|Q|}\sum_{j=1}^{|Q|} P@k_{j}$$

\paragraph{Mean Average Precision} Choosing a specific k to measure the Precision at k can be considered quite arbitrary. Average Precision addresses this by measuring precision at every possible threshold---for every relevant item, precision is measured up to and including the position of the item. Average Precision also has an implicit recall component because it accounts for relevant items \cite{buttcher2016information}. If the user interface with a specific number of items shown is being tested, then Average Precision is a more comprehensive measure compared to Precision at k.

$$AP = \frac{1}{|\text{Rel}|} \sum_{i=1}^{|\text{Res}|} \text{relevant}(i) \cdot P@i,$$

\noindent
where \textit{relevant(i)} is 1 if the i-th item in \textit{Res} is relevant; 0 otherwise. Similarly to $P@k$, $AP$ refers to the result of a single query.
We are, however, interested in the value of $AP$ aggregated over all queries, which is the Mean Average Precision denoted as $MAP$.

$$MAP = \frac{1}{|Q|}\sum_{j=1}^{|Q|} AP_{j}$$

\paragraph{Mean Reciprocal Rank} The user may be interested in one specific relevant item given their query. Reciprocal rank models that situation by favoring the rankings where the relevant document is as close to the top as possible \cite{buttcher2016information}.

$$RR = \frac{1}{\min\{k | \text{Res}[k] \in \text{Rel}\}}$$

\noindent
Accordingly, the Mean Reciprocal Rank (MRR) is computed as the average of the reciprocal ranks of the first relevant answers across all queries.

$$MRR = \frac{1}{|Q|}\sum_{j=1}^{|Q|} RR_{j}$$

\paragraph{Median Rank} Median Rank is calculated as the median of the ranks of the first relevant answers across all queries. In the presence of MRR, Median Rank may be considered redundant because, similarly to MRR, it favors rankings with relevant items near the top. We report this primarily for consistency with previous works \cite{hill2016learning, qi2020wantwords, zhang2020multi}. Also, when the number of possible results returned is capped, and there are no relevant items among the returned results, it is impossible to accurately estimate the rank of the first relevant item. Thus, an arbitrary value (1000 in this case) is chosen as exemplified by \citet{qi2020wantwords} in their project's GitHub repository.\footnote{\url{https://github.com/thunlp/WantWords}}

\paragraph{Accuracy@k} Top-k accuracy represents the proportion of responses to queries where at least one returned item is relevant regardless of its position. The expectation behind this metric is that the user will be satisfied if they see at least one relevant result in the output. Note that Accuracy@1 is equivalent to Mean Precision@1.

\begin{table*}[ht]
\centering
\begin{tabular}{lrrrrrrr}
\hline
Model                & MAP   & MP@1   & MP@10  & MRR   & Acc@1 & Acc@10 & Median Rank \\ \hline
E5 query-passage     & \textbf{0.4282} & \textbf{0.4952} & \textbf{0.1591} & \textbf{0.5470} & \textbf{0.4952} & \textbf{0.6438} & \textbf{1}           \\
E5 query-query       & 0.4202 & 0.4940 & 0.1571 & 0.5448 & 0.4940 & 0.6397 & \textbf{1}           \\
LaBSE                & 0.4081 & 0.4894 & 0.1502 & 0.5345 & 0.4894 & 0.6178 & \textbf{1}           \\
OpenAI               & 0.3746 & 0.4934 & 0.1376 & 0.5347 & 0.4934 & 0.6114 & \textbf{1}           \\
DistilUSE V2         & 0.3381 & 0.4544 & 0.1241 & 0.4894 & 0.4544 & 0.5526 & 2           \\
BGE                  & 0.2660 & 0.4323 & 0.1038 & 0.4607 & 0.4323 & 0.5090 & 7           \\
Word2Vec             & 0.2573 & 0.4203 & 0.1040 & 0.4518 & 0.4203 & 0.5072 & 8           \\
MPNet                & 0.2552 & 0.4255 & 0.0997 & 0.4516 & 0.4255 & 0.4954 & 11          \\
DistilUSE V1         & 0.2389 & 0.4048 & 0.0897 & 0.4247 & 0.4048 & 0.4569 & 74          \\
XLM-RoBERTa          & 0.2306 & 0.4065 & 0.0901 & 0.4281 & 0.4065 & 0.4637 & 51          \\  \hline

\end{tabular}
\caption{Reverse dictionary performance in unlabeled evaluation on the full Sõnaveeb dataset, with both Estonian and non-Estonian queries. The E5 model is evaluated in two settings: by using passage prefix for candidate definitions (E5 query-passage) and by using query prefix for candidate definitions (E5 query-query).}
\label{tab:mult_unlabeled}
\end{table*}

\begin{table*}[h]
\centering
\begin{tabular}{lrrrrrrr}
\hline
Model          & MAP   & MP@1   & MP@10  & MRR   & Acc@1 & Acc@10 & Median Rank \\ \hline
LaBSE & \textbf{0.3913} & 0.4546 & 0.1449 & 0.4738 & 0.4546 & 0.5122 & 6 \\
DistilUSE V1 & 0.3873 & 0.4635 & 0.1451 & 0.4827 & 0.4635 & 0.5215 & 4 \\
DistilUSE V2 & 0.3775 & 0.4367 & 0.1411 & 0.4561 & 0.4367 & 0.4951 & 12 \\
E5 query-passage & 0.3746 & \textbf{0.4732} & \textbf{0.1515} & \textbf{0.5005} & \textbf{0.4732} & \textbf{0.5540} & \textbf{2} \\
E5 query-query & 0.3627 & 0.4448 & 0.1458 & 0.4720 & 0.4448 & 0.5269 & 4 \\
OpenAI & 0.3335 & 0.4624 & 0.1447 & 0.4901 & 0.4624 & 0.5464 & \textbf{2} \\
BGE & 0.2404 & 0.4356 & 0.1173 & 0.4569 & 0.4356 & 0.4992 & 10 \\
MPNet & 0.2204 & 0.4053 & 0.1126 & 0.4259 & 0.4053 & 0.4675 & 32 \\
XLM-RoBERTa & 0.1525 & 0.3955 & 0.0822 & 0.4048 & 0.3955 & 0.4229 & 1000 \\
Word2Vec & 0.1326 & 0.3617 & 0.0732 & 0.3775 & 0.3617 & 0.4087 & 1000 \\ \hline
\end{tabular}
\caption{Reverse dictionary performance in unlabeled evaluation using only non-Estonian queries. The E5 model is evaluated in two settings: by using passage prefix for candidate definitions (E5 query-passage) and by using query prefix for candidate definitions (E5 query-query).}
\label{tab:mult_unlabeled_rest}
\end{table*}

\section{Results}

We expect that the output of a good reverse dictionary system would contain as many relevant items as possible and they would be as close to the top as possible. Meanwhile, the user's interest seems unlikely to be strictly limited to a single item. Therefore, although we show all measures described in Section~\ref{sec:metrics}, we consider Mean Average Precision (MAP) as the main measure because it rewards situations where relevant items are grouped at the top.

\paragraph{Unlabeled evaluation}
The main purpose of this evaluation was to measure the quality of different embedding models and rank them to select the best one when applied to creating a reverse dictionary system. 
We can single out three main aspects that can affect the position of the model on the table: 1) whether the target language was in the model's training data, 2) whether the model was trained for retrieval, and 3) whether the model was trained for cross-lingual retrieval specifically.

\begin{table*}[ht]
\centering
\begin{tabular}{llrrrrrrr}
\hline
 & Model            & MAP   & MP@1   & MP@10  & MRR   & Acc@1 & Acc@10 & Median Rank \\ \hline
\multicolumn{9}{c}{Definitions in Estonian} \\ \hline
& E5 query-query   & \textbf{0.2135} & \textbf{0.2600} & \textbf{0.1435} & \textbf{0.3635} & \textbf{0.2600} & \textbf{0.5700} & \textbf{6}           \\
& E5 query-passage & 0.2024 & 0.2400 & 0.1350 & 0.3425 & 0.2400 & 0.5350 & \textbf{6}           \\
& LaBSE            & 0.1432 & 0.2150 & 0.1055 & 0.2885 & 0.2150 & 0.4500 & 14          \\
& OpenAI           & 0.1031 & 0.1750 & 0.0835 & 0.2390 & 0.1750 & 0.3850 & 27          \\
& DistilUSE V2     & 0.0983 & 0.1400 & 0.0590 & 0.1979 & 0.1400 & 0.3000 & 49          \\ \hline

\multicolumn{9}{c}{Definitions in English} \\ \hline
& E5 query-query   & \textbf{0.194}8 & 0.2050 & \textbf{0.1250} & \textbf{0.3013} & 0.2050 & \textbf{0.5250} & \textbf{8}           \\
& E5 query-passage & 0.1717 & 0.2150 & 0.1135 & 0.2952 & 0.2150 & 0.4750 & 13          \\
& LaBSE            & 0.1478 & \textbf{0.2300} & 0.1185 & 0.3074 & \textbf{0.2300} & 0.4850 & 12   \\
& DistilUSE V2     & 0.1262 & 0.1700 & 0.0885 & 0.2328 & 0.1700 & 0.3550 & 29          \\
& OpenAI           & 0.0996 & 0.1150 & 0.0870 & 0.1981 & 0.1150 & 0.4000 & 17          \\ \hline

\multicolumn{9}{c}{Definitions in Russian} \\ \hline
& E5 query-query   & \textbf{0.1954} & \textbf{0.2150} & \textbf{0.1385} & \textbf{0.3149} & \textbf{0.2150} & \textbf{0.5500} & \textbf{7}           \\
& LaBSE            & 0.1446 & 0.2050 & 0.1140 & 0.2922 & 0.2050 & 0.4750 & 14          \\
& E5 query-passage & 0.1166 & 0.1500 & 0.0945 & 0.2236 & 0.1500 & 0.3950 & 23          \\
& DistilUSE V2     & 0.1023 & 0.1150 & 0.0800 & 0.1874 & 0.1150 & 0.3350 & 39  \\ 
& OpenAI           & 0.0093 & 0.0050 & 0.0055 & 0.0231 & 0.0050 & 0.0500 & 1000        \\ \hline

\end{tabular}
\caption{Reverse dictionary performance on a labeled dataset in both monolingual and cross-lingual setting. }
\label{tab:labeled_eval}
\end{table*}

 Table \ref{tab:mult_unlabeled} demonstrates the performance of all tested models. We observe that the models possessing all three aspects mentioned above (trained on the target language, trained for retrieval, and trained for cross-lingual retrieval) dominate the table: E5 and LaBSE were both trained for cross-lingual retrieval and contained Estonian in their training data. Meanwhile, we have no information on how OpenAI's embedding model was trained. We can infer that it is also a retrieval model that has seen Estonian; however it was possibly not trained for cross-lingual retrieval specifically. 
DistilUSE V2, which also included Estonian in the training data, performs better than V1, as expected. The English-only retrieval models (BGE and MPNet) perform similarly to the Estonian Word2Vec baseline. Surprisingly, the very simple Word2Vec embedding model outperforms the much larger multilingual XLM-RoBERTa, even though the latter also includes Estonian.

\paragraph{Cross-lingual unlabeled evaluation}
The Estonian language heavily dominates the data used for the unlabeled evaluation since it comes from the Estonian language resource. However, it also contains about 14K definitions in other languages.
The results in Table \ref{tab:mult_unlabeled_rest} focus specifically on the cross-lingual capabilities of the models: only definitions in languages other than Estonian were used as queries, while the candidates remained the Estonian words with definitions in all available languages. 
All measures are slightly worse in this setting, showing that the cross-lingual search is harder than the monolingual retrieval. The overall ranking of the models remains roughly the same, with LaBSE being the best while the DistilUSE models are in second place. Interestingly, the V1 model that did not include Estonian in the training data is slightly better than the V2 model. The OpenAI model performs considerably worse on MAP than the other multi-lingual retrieval models.

\paragraph{Labeled evaluation}

 To ensure the results output by the unlabeled environment are reliable, we tested the models on a smaller labeled dataset focusing on modeling cross-lingual retrieval. Table~\ref{tab:labeled_eval} shows the labeled evaluation results of the best models from the unlabeled evaluation setting. We can see that the order of the models based on the metrics is approximately the same, which confirms the reliability of the unlabeled evaluation. We also note the relatively poor performance of OpenAI's embedding model, which further points towards a lack of focus on cross-linguality during its training. 
The labeled evaluation measures are lower compared to the unlabeled evaluation setting. It might be because the definitions from the small labeled dataset are very out of distribution compared to the dictionary definitions, so it makes sense that there would be some discrepancy.

 \paragraph{Qualitative evaluation} We manually examined some outputs of the best-performing E5 model using the definitions from our unlabeled evaluation dataset with few examples shown in Table \ref{tab:exemples}. We note that while the expected target words do not always appear in top-10 results, the output is sensible, and the other words in the output very often fit the queried definitions (which are sometimes quite vague) quite well. The same observation also applies to the cross-lingual search environment.

\section{Conclusion}
We proposed an IR-based reverse dictionary system leveraging pre-trained transformer-based language models for semantic search and evaluated it thoroughly on an Estonian language dictionary resource, the Sõnaveeb. 
Unlike the prediction-based approaches that have been mainly focused on in recent works, the IR-based approach does not require training any specialized models. Furthermore, it can be easily adapted to work in the cross-lingual setting.
The evaluations using the unlabeled evaluation procedure based on the synonym relations of the target dictionary and a small multilingual dataset showed that language models trained for cross-lingual retrieval are optimal for our use case. 
We expect that the proposed approach to building and validating reverse dictionary systems is reusable, approachable, and generalizable due to its simplicity, thus facilitating the development and improvement of language resources for less-represented languages with a focus on language learning, documentation, and preservation.

In future work, we would like to develop a user interface to perform human evaluation of our proposed system. The ultimate goal is to integrate the best-performing model into the existing Sõnaveeb language portal to enrich this resource by making the search and navigation more accessible and to support language learners via cross-lingual search.

\section*{Limitations}

The main limitation of our work is the lack of human evaluation. Although both the unlabeled and labeled evaluation approach attempt to model user satisfaction,  we do not know the correlation between automatic measures and human judgments at this point.
Although we assumed that MAP is the most suitable measure, its correlation with human judgments for our task still needs to be established.

Another limitation is that we have not assessed the level of noise in the dictionary resource Sõnaveeb used in this work. It is possible that filtering out non-informative definitions or possibly erroneous synonymy relations could result in a more precise evaluation. However, the main point of our unlabeled validation approach was to facilitate the performance of the different embedding models for building an IR-based reverse dictionary and to make it require as little additional effort as possible. We expect that any additional data filtering does not result in any significant change to the model rankings. This assumption is also supported by the results we obtained using the labeled evaluation.

Finally, when we consider the fact that the ultimate purpose of our work is to enable reverse dictionary search functionality in an existing language resource that real users could utilize via a graphical user interface, the way the reverse dictionary output would be represented visually would also play a significant role in user satisfaction. Consequently, the effect the visual presentation has on the performance of a reverse dictionary system also needs to be evaluated to get the full picture. We leave that particular aspect for future work.

\section*{Acknowledgments}

This research was supported by the Estonian Research Council Grant PSG721.

\bibliography{custom}

\end{document}